%% file: main-camera-ready.tex
\documentclass[sigconf]{acmart}

\AtBeginDocument{%
  }

\copyrightyear{2026}
\acmYear{2026}
\setcopyright{cc}
\setcctype{by}
\acmConference[WWW Companion '26]{Companion Proceedings of the ACM Web Conference 2026}{April 13--17, 2026}{Dubai, United Arab Emirates}
\acmBooktitle{Companion Proceedings of the ACM Web Conference 2026 (WWW Companion '26), April 13--17, 2026, Dubai, United Arab Emirates}
\acmPrice{}
\acmDOI{10.1145/3774905.3793122}
\acmISBN{979-8-4007-2308-7/2026/04}

\begin{document}

\title[PaperDebugger]{PaperDebugger: A Plugin-Based Multi-Agent System for In-Editor Academic Writing, Review, and Editing}

\author{Junyi Hou}
\email{e0945797@u.nus.edu}
\orcid{0009-0003-0443-456X}
\affiliation{%
  \institution{National University of Singapore}
  \city{Singapore}
  \country{Singapore}
}

\author{Andre Lin Huikai}
\email{andre_lin@u.nus.edu}
\orcid{0009-0007-5941-8005}
\affiliation{%
  \institution{National University of Singapore}
  \city{Singapore}
  \country{Singapore}
}

\author{Nuo Chen}
\email{nuochen@u.nus.edu}
\orcid{0000-0001-6563-1215}
\affiliation{%
  \institution{National University of Singapore}
  \city{Singapore}
  \country{Singapore}
}

\author{Yiwei Gong}
\email{imwithye@gmail.com}
\orcid{0009-0008-5629-9538}
\affiliation{%
  \institution{Independent}
  \city{Singapore}
  \country{Singapore}
}

\author{Bingsheng He}
\email{dcsheb@nus.edu.sg}
\orcid{0000-0001-8618-4581}
\affiliation{%
  \institution{National University of Singapore}
  \city{Singapore}
  \country{Singapore}
}

\renewcommand{\shortauthors}{Hou et al.}

\begin{abstract}
  Large language models are increasingly embedded into academic writing workflows, yet existing assistants remain external to the editor, preventing deep interaction with document state, structure, and revision history. This separation makes it impossible to support agentic, context-aware operations directly within LaTeX editors such as Overleaf. We present \textbf{PaperDebugger}, an \textbf{in-editor}, \textbf{multi-agent}, and \textbf{plugin-based} academic writing assistant that brings LLM-driven reasoning directly into the writing environment. Enabling such in-editor interaction is technically non-trivial: it requires reliable bidirectional synchronization with the editor, fine-grained version control and patching, secure state management, multi-agent scheduling, and extensible communication with external tools. PaperDebugger addresses these challenges through a Chrome-approved extension, a Kubernetes-native orchestration layer, and a Model Context Protocol (MCP) toolchain that integrates literature search, reference lookup, document scoring, and revision pipelines. Our demo showcases a fully integrated workflow, including localized edits, structured reviews, parallel agent execution, and diff-based updates, encapsulated within a minimal-intrusion user interface (UI). Early aggregated analytics demonstrate active user engagement and validate the practicality of an editor-native, agentic writing assistant. More details about this demo and video could be found at \texttt{\url{https://github.com/PaperDebugger/PaperDebugger}}.
\end{abstract}





\maketitle

\section{Introduction}
\input{sections/introduction.tex}

\section{System Overview}
\input{sections/system-overview.tex}

\section{Usage Analytics}
\input{sections/usage-analytics.tex}

\section{Demonstration}
\label{sec:demo}
This section presents two example in-editor workflows enabled by
PaperDebugger. Conference attendees may experiment with additional
workflows by installing the PaperDebugger extension directly from
the Chrome Web Store and applying it to their own Overleaf
projects.

\subsection{In-Editor Editing and Patch}
\label{subsec:critique_case}
\begin{figure}[t]
    \centering
    \includegraphics[width=0.99\linewidth]{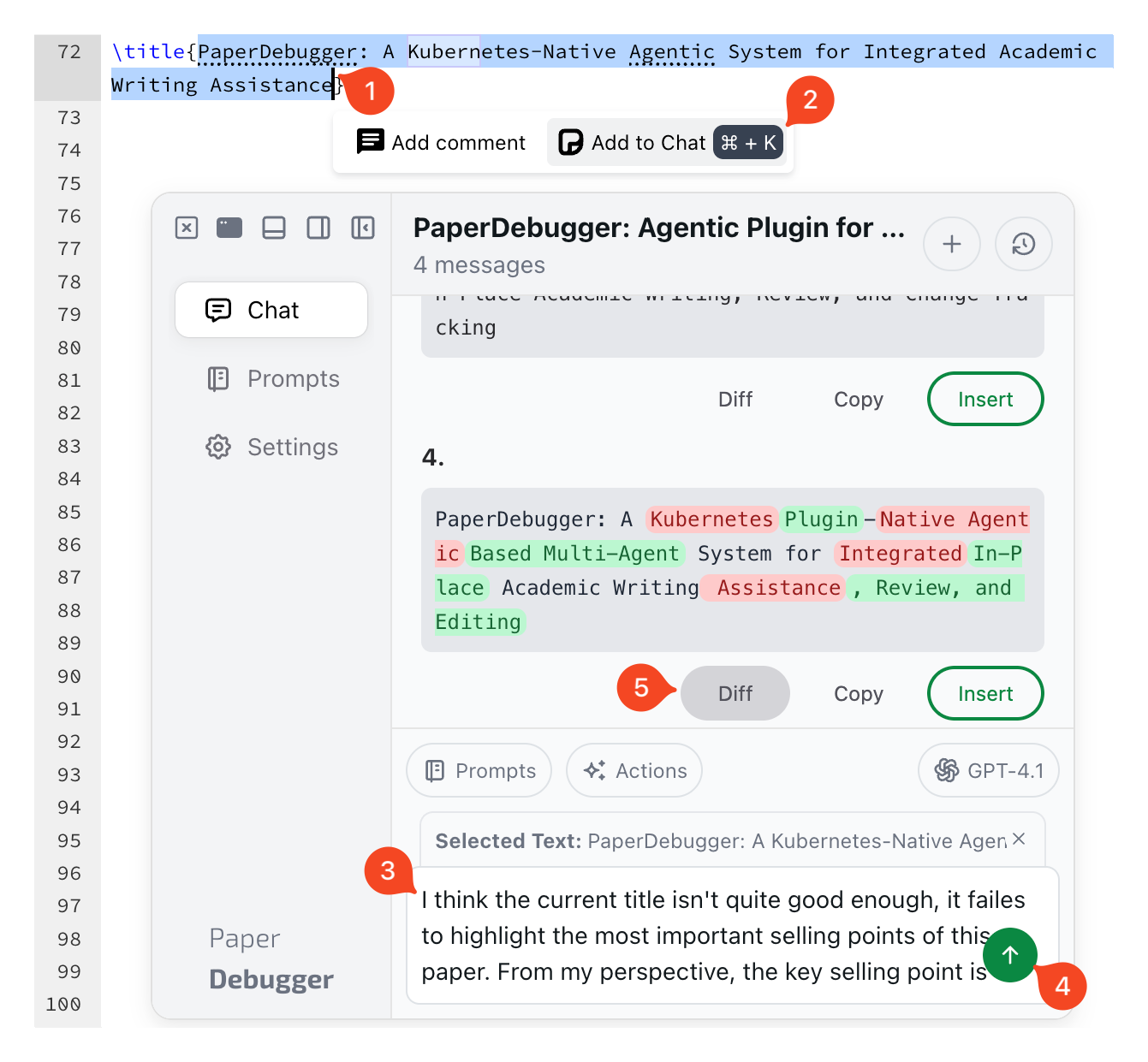} 
    \caption{In-editor editing workflow in PaperDebugger. (1) Select a span of LaTeX in Overleaf. (2) Add the selection to the PaperDebugger panel. (3) Specify the critique request. (4) Trigger the agentic pipeline. (5) Review and apply the returned before-after patches.}
    \Description{In-editor editing workflow in PaperDebugger. (1) Select a span of LaTeX in Overleaf. (2) Add the selection to the PaperDebugger panel. (3) Specify the critique request. (4) Trigger the agentic pipeline. (5) Review and apply the returned before--after patches.}
    \label{fig:frontend-workflow}
\end{figure}

A common task in AI-assisted academic writing is polishing or
refining text with AI agents. For example, when revising a conference
submission, an author may encounter an unclear section title and
highlight it directly within Overleaf to request a structured
rewrite. As shown in Figure~\ref{fig:frontend-workflow}, the
selected text is forwarded to the PaperDebugger panel, where the
author initiates a critique or refinement request.

Once invoked, PaperDebugger runs a coordinated pipeline with critique agent, enhancer, and patch generation agents, returning before-after
diffs that can be inspected and applied directly within the editor.

Candidate patches appear as inline previews, allowing the author to review the rationale behind each suggestion, selects the
preferred revision, and applies it with a single click. This workflow
replaces traditional copy-paste interactions with a seamless, context-aware
editing loop inside Overleaf.

Overall, this case demonstrates how PaperDebugger transforms a
simple editing action into a transparent and agentic patch workflow
that enhances clarity and structure while preserving an efficient,
in-editor writing experience.

\subsection{Deep Research and Comparative Analysis}

\label{subsec:research_case}

Another common task in academic writing is preparing the
related-work section of a manuscript, which requires understanding
how the current contribution compares with recent literature. The
author highlights the section header and requests ``deep research.''
PaperDebugger invokes an MCP-powered retrieval pipeline that
performs multi-stage semantic search over arXiv and curated
corpora. Within seconds, the system returns a ranked list of relevant
papers enriched with metadata, abstracts, and LLM-generated
explanations of relevance.

Upon selecting a target paper, the author activates \textit{Compare My
Work}. PaperDebugger automatically extracts key aspects---goals,
datasets, methods, evaluation protocols, and limitations---from both
papers and produces a structured side-by-side comparison. The
system highlights conceptual overlaps, methodological differences,
and missing dimensions in the author's draft. A citation-ready
summary table is also generated for direct insertion into the
manuscript.

For broader situational awareness, the author may conduct an
additional search (e.g., ``find relevant papers to read'').
PaperDebugger aggregates multiple related works into a consolidated
research map, revealing clusters such as mitigation-focused
systems, benchmarking frameworks, and hybrid evaluation pipelines.
The system further generates ``takeaways for positioning your
work'' that help the author articulate how their manuscript fits into
the broader landscape.

All outputs of the deep research workflow are delivered directly
in-editor, eliminating the need for manual copying of abstracts,
comparison tables, or summaries. This scenario demonstrates the
emerging capability of research-level reasoning and literature
synthesis directly within the writing environment.

\begin{figure}[t]
    \centering
    \includegraphics[width=0.99\linewidth]{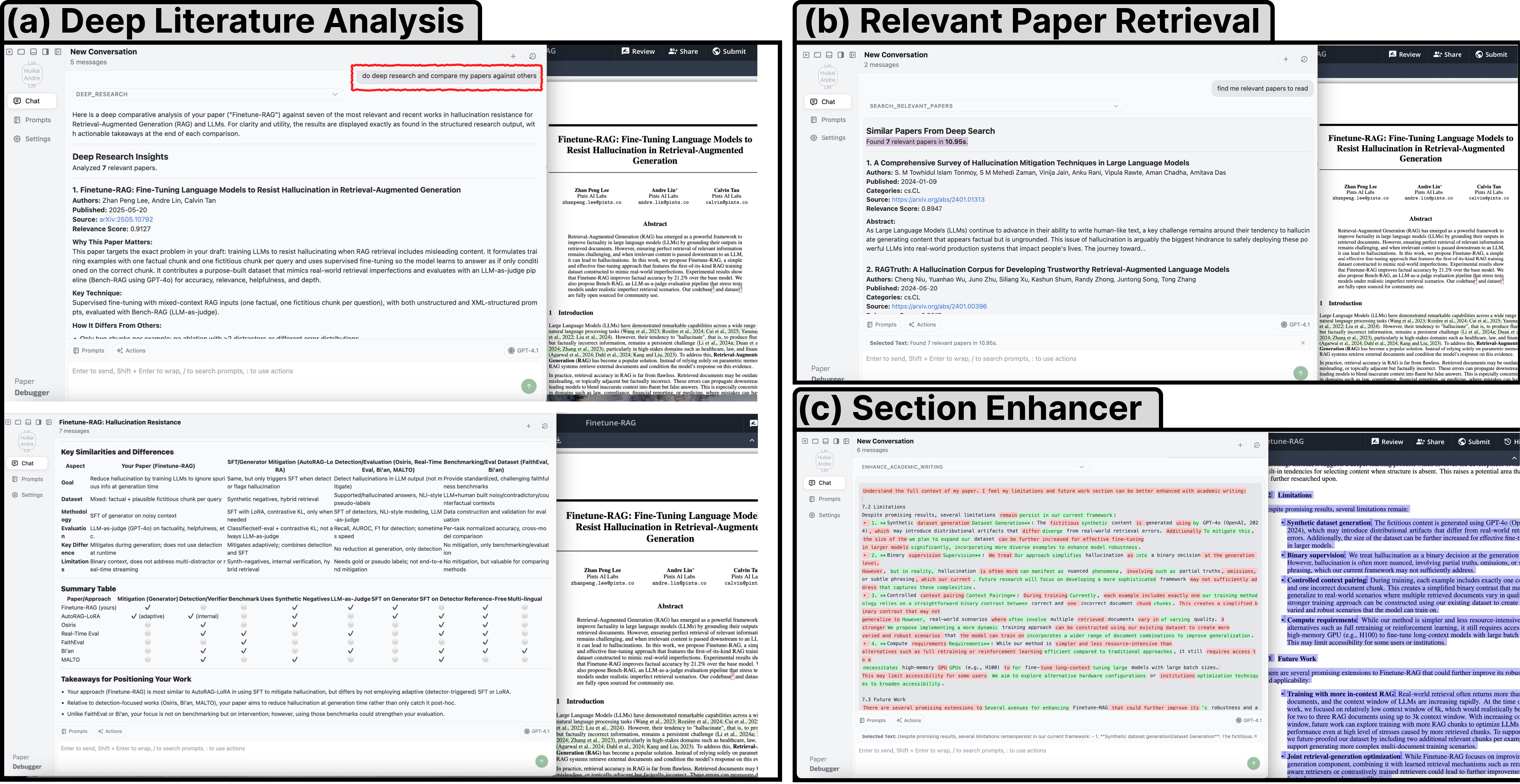}
    \caption{Example of an end-to-end research-use scenario supported by PaperDebugger. The system integrates XtraMCP (a) deep research, (b) related-paper retrieval, and (c) section enhancer to help authors understand, compare, and refine academic content within the editor.}
    \Description{Example of an end-to-end research-use scenario supported by PaperDebugger. The system integrates XtraMCP (a) deep research, (b) related-paper retrieval, and (c) section enhancer to help authors understand, compare, and refine academic content within the editor.}
    \label{fig:research-case}
\end{figure}



\section{Conclusion}

PaperDebugger provides a unified, in-editor academic writing environment that closes the long-standing gap between document editors and LLM-assisted workflows. By integrating directly with Overleaf through a Chrome-approved extension, the system enables context-aware critique, structured review, literature retrieval, and deterministic diff-based editing. Our real-world deployment demonstrates both feasibility and impact. Early usage analytics show sustained engagement across real writing projects, while case studies highlight how PaperDebugger supports both micro-level refinement and deep research tasks. The system is fully available through the Chrome Web Store, and conference attendees can experience the workflows firsthand by installing the extension and applying it to their own Overleaf documents.

\textbf{AI Usage Statements:} Portions of this manuscript were polished using PaperDebugger, but no scientific content was generated by AI. A complete policy statement and usage disclaimer are available in our project repository.

\begin{acks}
This research is supported by the National Research Foundation, Singapore and Infocomm Media Development Authority under its Trust Tech Funding Initiative. Any opinions, findings and conclusions or recommendations expressed in this material are those of the authors and do not reflect the views of National Research Foundation, Singapore and Infocomm Media Development Authority.

\end{acks}

\bibliographystyle{ACM-Reference-Format}
\bibliography{bibfile}

\end{document}

%% file: sections/introduction.tex
Large language models (LLMs) are increasingly used in assisting academic writing workflows, from brainstorming and outlining to line-level editing and reviewer-response drafting. Recent systems for human–AI co-writing and assisted feedback demonstrate that LLM-based suggestions can improve fluency and reduce mechanical writing effort at scale~\cite{lee2024designspace,liebling2025towards}. Research on writing-support tools further shows that structured interventions can meaningfully improve writing efficiency and user experience~\cite{sarrafzadeh2020stageaware,mysore2024pearl,ingley2023leveraging}.

Despite these developments, the majority of existing tools still operate outside the primary editing environment, requiring copy-and-paste workflows and fragmenting interaction history~\cite{wen2024overleafcopilot,lee2024designspace}. This external workflow introduces context switching, breaks writing flow, and makes revision history difficult to preserve. Additionally, external tools provide limited revision provenance; feedback, applied changes, and reasoning disappear once the interaction window closes. Tools such as Writefull provide in-editor language suggestions but remain largely surface-level, offering limited transparency into applied changes~\cite{writefull}.

To address these challenges, we present \textbf{PaperDebugger}, an in-editor LLM agent system that integrates directly into Overleaf, a widely used academic writing editor. Instead of treating the writing process and model interaction as separate workflows, the system enables critique, refinement, and revision to take place in context, inline, and tied to document structure. The system provides persistent interaction history, patch-based edits, and structure-aware feedback while preserving the continuity of writing. Technically, it is implemented as a Chrome extension that communicates with a Kubernetes-based backend using gRPC. The Model Context Protocol (MCP) acts as a lightweight extensibility layer, enabling modular functionality and future agent capabilities without altering the core architecture. PaperDebugger is fully implemented with over 24,000 lines of code.

The current implementation of PaperDebugger has been deployed to real users via the Chrome Web Store~\footnote{\url{https://chromewebstore.google.com/detail/paperdebugger/dfkedikhakpapbfcnbpmfhpklndgiaog}} and used in live academic writing scenarios. Our demo showcases a complete workflow: authors open a \LaTeX{} project, activate PaperDebugger to request critiques for selected passages, inspect proposed revisions in a diff-style view, and apply accepted patches back into the editor with a single click. In addition to revision workflows, PaperDebugger allows users to invoke MCP-based tools, such as related literature retrieval, directly from the editor, thereby facilitating seamless insertion of relevant, high-quality references. Early analytics based on anonymized telemetry indicate sustained user engagement and active adoption, demonstrating both the technical feasibility and practical value of an in-editor, agentic writing assistant.

To summarize, PaperDebugger makes three key contributions:

\begin{itemize}
    \item \textbf{In-editor academic writing assistance} that integrates directly with Overleaf and operates on selected text, eliminating copy--paste workflows and preserving full writing flow and document context.
    \item \textbf{A scalable multi-agent execution architecture} implemented through Kubernetes-driven pod orchestration, enabling parallel reasoning, structured review, MCP-powered retrieval, AI reviewer, and deterministic diff-based editing.
    \item \textbf{Evidence of real-world usability and adoption}, supported by deployment through the Chrome Web Store and early anonymized telemetry demonstrating repeated use of critique and revision workflows in authentic writing settings.
\end{itemize}

%% file: sections/system-overview.tex

\subsection{Architecture Overview}
\begin{figure}[h]
	\centering
	\includegraphics[width=0.95\linewidth]{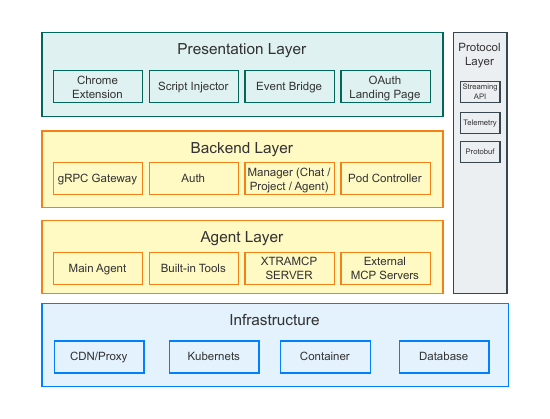}
	\caption{Overall architecture of PaperDebugger, consisting of the presentation layer, backend layer, agent layer, protocol layer and infrastructure.}
    \Description{Architecture diagram showing five layers: presentation, backend, agent, protocol, and infrastructure.}
	\label{fig:architecture}
\end{figure}

As shown in Figure~\ref{fig:architecture}, PaperDebugger consists of five layers: (1) a presentation layer integrated directly into Overleaf, (2) a backend layer that manages workflow execution, (3) an agent layer running containerized tools, (4) a protocol layer handling structured communication, (5) and an infrastructure layer providing storage and operational services. The presentation layer is implemented as a Chrome extension that injects UI components into Overleaf and synchronizes project context and user actions. The backend layer coordinates authentication, session state, and workflow routing, exposing a streaming interface through a gRPC gateway.


\textbf{Presentation Layer.} PaperDebugger integrates directly into Overleaf via a lightweight Chrome extension. The extension injects a floating panel and inline action buttons next to highlighted \LaTeX{} spans. When users trigger a workflow, the extension captures the selected text and project state, then communicates with the backend using streamable gRPC. Edits are presented as before–after diffs, and accepted patches are applied instantly to the \LaTeX{} source. This eliminates copy–paste cycles, maintains consistent revision history, and creates a seamless user experience.

\textbf{Protocol Layer.} Communication between the chrome extension and backend leverages a custom message streaming protocol, compatible with OpenAI’s server-sent event (SSE) format. This protocol supports real-time streaming of intermediate model outputs, allowing users to receive updates during multi-step workflows.

\textbf{Backend Layer.} The backend is implemented in Go and deployed on Kubernetes. It orchestrates stateless LLM agents, each running inside isolated pods, enabling high concurrency and horizontal scaling. The orchestrator handles routing, model selection, permission checks, and schema validation.

\textbf{Agent Layer.} The agent layer supports two execution modes: \emph{prompt-template agents} and \emph{workflow-based agents}. Prompt-template agents are lightweight, single-shot LLM invocations defined by structured templates. They are designed for low-latency tasks such as grammar polishing. Workflow-based agents are declarative workflows that coordinate multiple LLM calls, tool executions, and validation steps. These workflows handle complex tasks like deep research, relevant paper retrieval, and full-document enhancement.

\subsection{Agentic Design}
Building on these two execution modes, PaperDebugger extends beyond a single-model rewriting tool through the XtraMCP architecture, a refined variant of MCP tailored for academic writing. XtraMCP exposes a suite of validated tools for literature search, affiliation lookup, and structured data extraction, and enforces our internal Pydantic-based schemas and internal consistency checks to minimize hallucinations. Concretely, three core MCP-powered components back the agents: (i) a low-latency embedding + LLM re-ranking pipeline that provides high-quality semantic retrieval and real-time literature lookup; (ii) a multi-step AI review pipeline, inspired by conference reviewing workflows like AAAI, that guides the Reviewer agent through targeted, segment-level critique; and (iii) XtraGPT~\cite{chen2025xtragptcontextawarecontrollableacademic}. XtraGPT is a model suite tuned for academic writing, ensuring that suggested revisions are context-aware, properly scoped, and phrased in appropriate scholarly style, more details can be found in the paper~\cite{chen2025xtragptcontextawarecontrollableacademic}.

On top of this foundation, PaperDebugger runs a suite of specialized agents: a \emph{Reviewer} agent that produces structured critique, an \emph{Enhancer} agent for rewriting and refinement, a \emph{Scoring} agent for clarity and coherence evaluation, and a \emph{Researcher} agent that performs literature lookup via XtraMCP tools. For full-document review requests, a coordinating agent decomposes the task into segment-level sub-queries, dispatches them across worker pods, and merges the results into a unified output.

\begin{figure}[h]
    \centering
    \includegraphics[width=1\linewidth]{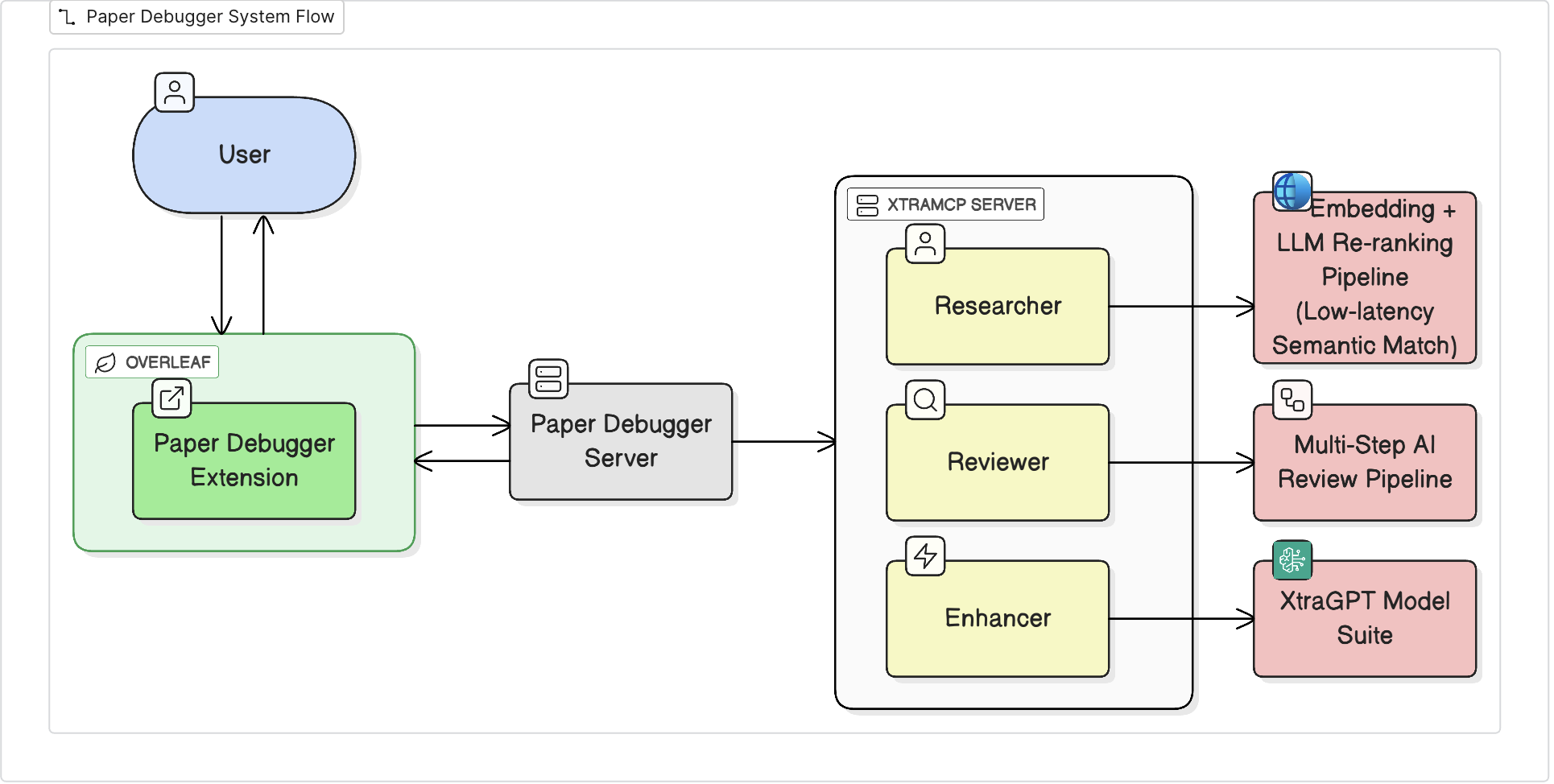}
    \caption{PaperDebugger end-to-end workflow: The extension captures user actions and sends them to the PaperDebugger server, which coordinates built-in agents or specialized agents on the XtraMCP server.}
    \Description{PaperDebugger end-to-end workflow: The extension captures user actions and sends them to the PaperDebugger server, which coordinates built-in agents or specialized agents on the XtraMCP server.}
    \label{fig:workflow}
    {\vspace{-2mm}}
\end{figure}

Figure~\ref{fig:workflow} illustrates the execution flow, showing how user actions are routed through the orchestration layer, activate the appropriate agents, and return deterministic diff-based edits back to the editor. We carefully design and validate each agent workflow, and the full prompt templates and agent specifications are available in our project repository.

%% file: sections/usage-analytics.tex
We analyze anonymized telemetry from both the Chrome extension and backend between May 2025 -- January 2026 to understand how PaperDebugger is used in real writing environments. In the following, we present the key statistics, and more usage analysis is in our project repository.

\paragraph{\textbf{Real-World Adoption}} Table \ref{tab:adoption-summary} shows early usage signals. With 4116 extension installs, 2761 registered users, and 732 monthly active users. Users created 3106 projects and 7447 writing threads, indicating sustained multi-session usage rather than one-off experimentation. User reviews reflect strong satisfaction with the integrated workflow (``convenience'', ``seamless''). At the same time, feedback highlights natural research value and limitations, such as domain-specific tone (“suggestions feel CS-like”) and performance drops on long documents.

\begin{table}[h]
\centering
\small
\caption{Early adoption metrics demonstrating real deployment and recurring use.}
\label{tab:adoption-summary}
\begin{tabular}{l r}
\toprule
\textbf{Metric} & \textbf{Value} \\
\midrule
Chrome extension installs & 4116 \\
Registered users & 2761 \\
Active users (30-day) & 732 \\
Projects created (all-time) & 3106 \\
Threads created (all-time) & 7447 \\
Chrome store user rating & 4.82 / 5 \\
\bottomrule
\end{tabular}
\end{table}

\paragraph{\textbf{Interaction Patterns}} Telemetry reveals heavy usage of in-editor revision rather than one-shot generation. Table~\ref{tab:interactions} highlights the three most frequent refinement actions:

\begin{table}[h]
\centering
\small
\caption{Most frequent in-editor refinement operations.}
\label{tab:interactions}
\begin{tabular}{l r}
\toprule
\textbf{Event Type} & \textbf{Count} \\
\midrule
Diff viewed & 4593 \\
Copy suggestion & 3214 \\
Insert patch & 2697 \\
\bottomrule
\end{tabular}
\end{table}

Three usage patterns stand out:

\begin{itemize}
    \item \textbf{Users iterate rather than one-shot.} Refinement actions recur within the same session.
    \item \textbf{Patch diffs are the dominant control surface.} Users frequently inspect diffs before applying changes.
    \item \textbf{Sessions contain multiple refinement events.} Interaction density indicates sustained, in-editor revision activity.
\end{itemize}